\documentclass[letterpaper, 10 pt, journal, twoside]{IEEEtran}

\IEEEoverridecommandlockouts                              

\title{Versatile LiDAR-Inertial Odometry with SE(2) Constraints for Ground Vehicles
}

\author{Jiaying Chen$^{1}$, Han Wang$^{1}$, Minghui Hu$^{1}$, and  Ponnuthurai Nagaratnam Suganthan$^{2}$ 
\thanks{Manuscript received: January, 9, 2023; Revised March, 11, 2023; Accepted April, 2, 2023.}
\thanks{This paper was recommended for publication by Editor Javier Civera upon evaluation of the Associate Editor and Reviewers' comments.
This work was supported by MIND POINTEYE PTE. LTD.} 
\thanks{$^{1}$ Jiaying Chen, Hang Wang, and Minghui Hu are with the School of Electrical and Electronic Engineering, Nanyang Technological University, Singapore 639798, 50 Nanyang Avenue, e-mail: {\tt\footnotesize  \{jiaying001, hwang027, e200008\}@e.ntu.edu.sg};}%
\thanks{$^{2}$ P. N. Suganthan is with the School of Electrical and Electronics Engineering, Nanyang Technological University, Singapore and KINDI Center for Computing Research, College of Engineering, Qatar University, Doha, Qatar, email: {\tt\footnotesize epnsugan@ntu.edu.sg}.}%
\thanks{Digital Object Identifier (DOI): see top of this page.}
}

\usepackage{amsmath}
\usepackage{xcolor}
\usepackage{adjustbox}
\usepackage{bm}
\usepackage{amsfonts}
\usepackage{graphicx}
\usepackage{stfloats}
\usepackage{float}
\usepackage{subcaption}
\usepackage{comment}
\usepackage{multirow}
\usepackage{hyperref}
\usepackage[ruled,vlined]{algorithm2e}

\newcommand\scalemath[2]{\scalebox{#1}{\mbox{\ensuremath{\displaystyle #2}}}}

\begin{document}

\maketitle

\begin{abstract}
LiDAR SLAM has become one of the major localization systems for ground vehicles since LiDAR Odometry And Mapping (LOAM). Many extension works on LOAM mainly leverage one specific constraint to improve the performance, \textit{e.g.}, information from on-board sensors such as loop closure and inertial state; prior conditions such as ground level and motion dynamics. In many robotic applications, these conditions are often known partially, hence a SLAM system can be a comprehensive problem due to the existence of numerous constraints. Therefore, we can achieve a better SLAM result by fusing them properly. In this paper, we propose a hybrid LiDAR-inertial SLAM framework that leverages both the on-board perception system and prior information such as motion dynamics to improve localization performance. In particular, we consider the case for ground vehicles, which are commonly used for autonomous driving and warehouse logistics. We present a computationally efficient LiDAR-inertial odometry method that directly parameterizes ground vehicle poses on SE(2). The out-of-SE(2) motion perturbations are not neglected but incorporated into an integrated noise term of a novel SE(2)-constraints model. For odometric measurement processing, we propose a versatile, tightly coupled LiDAR-inertial odometry to achieve better pose estimation than traditional LiDAR odometry.
Thorough experiments are performed to evaluate our proposed method's performance in different scenarios, including localization for both indoor and outdoor environments. The proposed method achieves superior performance in accuracy and robustness.
\end{abstract}

\begin{IEEEkeywords}
SLAM, localization, mapping, sensor fusion
\end{IEEEkeywords}
\section{INTRODUCTION} \label{section:INTRODUCTION} 
As one of the most fundamental research topics in robotics, Simultaneous Localization And Mapping (SLAM) has developed rapidly in recent decades. It is a task to build a dense 3-dimension map of an unknown environment and simultaneously localize the robot in the map based on onboard perception systems such as cameras, LiDAR, etc. Compared with the vision-based SLAM system \cite{forster2016svo,qin2018vins,campos2021orb}, a LiDAR sensor can obtain direct, dense, and accurate depth measurements of environments and is insusceptible from illumination and weather changes \cite{debeunne2020review}. Hence LiDAR-based SLAM systems are more reliable and precise. Over the last decade, LiDAR-based SLAM systems have been widely applied in many autonomous robot fields, such as autonomous driving \cite{levinson2011towards} and UAVs \cite{liu2017planning}.

The most classic LiDAR SLAM is LiDAR Odometry And Mapping (LOAM) \cite{zhang2014loam}, which provides a unified LiDAR-based localization framework in an unknown environment. However, LOAM is limited by its localization accuracy and computational efficiency, \textit{e.g.}, the attitude estimation is less accurate during large rotation. Fortunately, in many robotic applications, the environment can be partially known so that the prior conditions can be assumed to improve the performance. Therefore, the LiDAR SLAM is further extended recently. \cite{jiao2021greedy} and \cite{duan2022pfilter} enhance the feature extraction method in LOAM by actively selecting or filtering features. Moving objects often make the localization unreliable, therefore some works \cite{pfreundschuh2021dynamic,qian2022rf} are proposed  to preliminarily remove the moving points in the environment.
Moreover, with the advancement of the perception system, LiDAR-inertial is introduced with an external IMU unit, which has better robustness in detecting angle change, e.g., LIO-SAM \cite{shan2020lio}.

For a ground vehicle, it is often moving in 2D space so that the 6-DoF pose estimation can be approximated as SE(2) to represent the robot pose. A quick solution is to apply stochastic-SE(2) constraints on SE(3) poses \cite{wu2017vins,zheng2018odometry} by using the planar motion constraints for ground vehicle application. However, due to rough terrain or motion vibration, ground vehicle motion in real-world environments frequently deviates from the SE(2) constraints model. Thus, such an assumption of SE(2) motion is not robust to disturbance. To tackle the above problems, we present a 3D LiDAR-based SLAM system that solves the inevitable non-SE(2) perturbations and provides a practical real-time LiDAR-based SLAM solution to the public. We build a SE(2)-constrained pose estimation method for ground vehicles, which extends our previous work~\cite{wang2021f}. It can achieve real-time performance up to 20 Hz on a low-power embedded computing unit. The main contributions of this paper are summarised as follows:
\begin{itemize}
\item We propose a versatile LiDAR-inertial SLAM framework for the ground robot, which considered multiple common constraints to improve the localization accuracy. The implementation of our approach is open-sourced \footnote{\url{https://github.com/jiaying001/SE2LIO}}, and the accompanying video is available at \url{https://www.youtube.com/watch?v=SLfmFiB5aIk}.
\item We propose a SE(2) constraints model that incorporates the out-of-SE(2) perturbations, as well as planar and edge feature points from 3D LiDAR measurements.
\item A tightly coupled LiDAR-IMU odometry algorithm is proposed, which provides real-time accurate state estimation with a high update rate.
\item The proposed method achieves real-time performance up to 20 Hz on a low-power embedded computing unit. A thorough evaluation of the proposed method is presented, implying that our method is able to provide reliable and computationally efficient localization for ground robots.
\end{itemize}

This paper is organized as follows: Section \ref{section:Related work} presents a review of the related work on existing LiDAR-based SLAM approaches. Section \ref{section:Methodology} describes the details of the proposed approach, including IMU preintegration, feature point selection, laser points alignment, odometry estimation, the novel proposed SE(2) constraints, and mapping. Section \ref{section:EXPERIMENT EVALUATION} shows experiment results and analysis, followed by the conclusion in Section \ref{section:Conclusion}.

\section{Related work} \label{section:Related work} 

Prior information is often used to improve the performance of LiDAR SLAM. For feature point pairs matching, LOAM \cite{zhang2014loam} proposed a real-time SLAM solution. LOAM extracts edge and planar feature points from the raw point cloud and then minimizes the distance of point-to-line and point-to-plane. LOAM is a milestone in terms of efficiency and has sparked many other works \cite{shan2018lego,lin2020loam}. LeGO-LOAM \cite{shan2018lego} introduces ground segmentation and isotropic edge points extraction. Then a two-step non-linear optimization is executed on ground and edge correspondences to estimate the transformation. 

The fusion of perception systems is always an efficient and effective way to introduce more sensory constraints to improve performance. For example, incorporating an Inertial Measurement Unit (IMU) sensor can considerably increase the accuracy and robustness of LiDAR odometry. The IMU can measure robot motion at high frequency, which can bridge the gap between two consecutive LiDAR frames. The fusion scheme consists of two major categories: loosely-coupled fusion and tightly-coupled fusion. In LOAM \cite{zhang2014loam} and LeGO-LOAM \cite{shan2018lego}, the IMU is only introduced to de-skew compensates for the motion distortion in a LiDAR scan, which is not involved in the optimization process. The loosely-coupled method partially solves the rotation issue. Recently, there has been a growing focus on tightly-coupled
fusion. It was first introduced in \cite{lupton2011visual}, which parameterizes the rotation error using Euler angles, and  further improved in \cite{forster2016manifold,forster2015imu} by adding posterior IMU bias correction. LIO-Mapping \cite{ye2019tightly} introduces a tightly-coupled LiDAR-inertial fusion method, where the IMU preintegration is introduced into the odometry. However, LIO-Mapping cannot achieve real-time performance. Then, LIO-SAM \cite{shan2020lio} proposes a LiDAR-inertial odometry framework for real-time state estimation. 

LiDAR measurements are typically treated as a zero-mean Gaussian noise distribution \cite{thrun2002probabilistic}. However, the results of LiDAR odometry are prone to drift upward along the z-axis. 
For this problem, \cite{laconte2019lidar} presents a reasonable explanation that LiDAR measurement bias exists and could be up to 20 cm for the high incidence angle when the LiDAR scans the surface of the road at a far distance. As a result, the points observed from the ground are slightly bent, and the trajectory of LiDAR odometry drifts along the vertical direction. In \cite{laconte2019lidar}, a mathematical model is proposed to remove this bias. However, 
the bias is affected by other factors such as temperature, which makes this method hard to apply. Therefore, we leverage the ground constraint as prior information to reduce the effect of LiDAR measurement bias.
 
For ground vehicles, the SE(2) poses are often used to represent the robots' motion; 
for example,  Andor \cite{hess2016real} uses 2-D pose parameterization, which contains a 2-D translation vector and an orientation angle.
However, the vehicle motion in real-world environments is often out-of-SE(2) pose. To leverage the planar motion constraint effectively, stochastic SE(2)-constraints have been proposed in some visual-based localization systems \cite{wu2017vins,zheng2018odometry}. In LeGO-LOAM \cite{shan2018lego}, ground points are extracted and aligned to estimate 3 DOF of 6 DOF pose. However, they haven't fused ground information into the pose graph optimization framework. In this paper, we present that the vehicle pose is directly parameterized by SE(2) parameterization without neglecting the out-of-SE(2) motion perturbations by incorporating an integrated noise term into a novel SE(2)-constraints model. 
Compared to the stochastic SE(2)-constraints, our proposed constraints model is robust and simpler.

\begin{figure}[!t]
\centering
\includegraphics[width=0.99\linewidth]{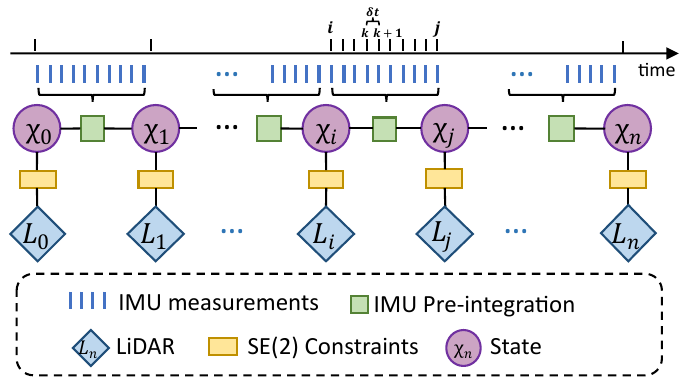}
\caption{The system structure of SE2LIO}
\label{fig: overview}
\end{figure}

\section{Methodology}\label{section:Methodology} 
\subsection{Problem Statement and System Overview}
We regard the odometry process as a SLAM problem, the robot state vector $\mathcal{X}$ is defined as:
\begin{equation}
    \label{robot state}
    \mathcal{X} =[\bm{\phi}^T, \mathbf{P}^T, \mathbf{V}^T, (\mathbf{b}^a)^T, (\mathbf{b}^g)^T]^T,
\end{equation}
where $\phi \in \mathfrak{so}(3)$ \cite{kirillov2008introduction}. The rotation matrix $\mathbf{R}$ on manifold $SO(3)$ is defined as $\exp{[ \boldsymbol{\phi}}]^\wedge$. $(\cdot)^\wedge$ denotes a skew-symmetric matrix. The transformation matrix $\mathbf{T}=[\mathbf{R} | \mathbf{P} ]$ from the robot body frame $\boldsymbol{B}$ to the world frame $\boldsymbol{W}$, while $\mathbf{V}$, $\mathbf{b^a}$ and $\mathbf{b^g}$ are velocity, accelerometer bias and gyroscope bias respectively. We assume that the IMU frame and LiDAR frame coincide with the robot body frame.

The overview of the proposed system is shown in Fig. \ref{fig: overview}. The system receives sensor data from a 3D LiDAR and an IMU. The aim of the system is to determine the state of the robot based on the measurements gathered from the LiDAR and IMU sensors, which is a Maximum A Posteriori (MAP) estimation problem \cite{barfoot2017state}. 

\subsection{Preintegrated IMU Constraints}
\label{Preintegrated IMU Constraints}
The IMU measures angular velocity $\hat{\mathbf{w}}_t$ and acceleration $\hat{\mathbf{a}}_t$ in $\boldsymbol{B}$, which are affected by the addictive white noise and a slowly varying sensor bias:
\begin{equation}
\label{imu_measurements}
\begin{split}
       \hat{\mathbf{w}}_t &=\mathbf{w}_t+\mathbf{b}_t^g+\mathbf{n}_t^g\\
  \hat{\mathbf{a}}_t &=\mathbf{R}_t^{\mathbf{BW}}(\mathbf{a}_t-\mathbf{g})+\mathbf{b}_t^a+\mathbf{n}_t^a,
\end{split} 
\end{equation}
where $\mathbf{n}^a$ and $\mathbf{n}^g$ represent measurement noise of the accelerometer and the gyroscope. 
$\mathbf{R}^{\mathbf{BW}}$ is the rotation matrix from $\mathbf{W}$ to $\mathbf{B}$. 
$\mathbf{g}$ is the constant gravity vector in $\mathbf{W}$.

We can use IMU measurements to deduce the robot's motion. The rotation, position, and velocity between time $k$ and $k+1$ can be obtained as:

\begin{equation}
\label{imu1}
\begin{split}
\exp{([\boldsymbol{\phi}_{i,k+1}]^\wedge)}& = \exp{([ \boldsymbol{\phi}_{i,k}]^\wedge)}\exp{([\bar{\mathbf{w}}\delta t]^\wedge)}\\
\mathbf{P}_{i,k+1}& =\mathbf{P}_{i,k}+  \mathbf{V}_{i,k}\delta t+ \frac{1}{2}\bar{\mathbf{a}}\delta t^2\\
\mathbf{V}_{i,k+1}& = \mathbf{V}_{i,k}+ \bar{\mathbf{a}}\delta t , 
\end{split}
\end{equation}   
where $\bar{\mathbf{w}}=\frac{1}{2}[(\hat{\mathbf{w}}_k-\mathbf{b}_k^g-\mathbf{n}_k^g)+(\hat{\mathbf{w}}_{k+1}-\mathbf{b}_{k+1}^g-\mathbf{n}_{k+1}^g)]$ and $\bar{\mathbf{a}}=\frac{1}{2}[ \mathbf{R}_{i,k}^\mathbf{WB}(\hat{\mathbf{a}}_k-\mathbf{b}_k^a-\mathbf{n}_k^a)+ \mathbf{R}_{i,k+1}^\mathbf{WB}(\hat{\mathbf{a}}_{k+1}-\mathbf{b}_{k+1}^a-\mathbf{n}_{k+1}^a)]+\mathbf{g}$. We use $\bar{w}$ and $\bar{a}$, the mid-point values of the instantaneous angular velocity and acceleration between two consecutive IMU measurements to increase the robustness.

Then, we apply the IMU preintegration method \cite{forster2016manifold} by iterating the IMU integration for all $\delta t$ intervals between
two consecutive LiDAR sweeps at timestamps $i$ and $j$. The relative robot motion in $\mathbf{B}$ can be obtained as:
\begin{equation}
\label{correction}
\scalemath{0.87}{
\begin{split}
 \exp{([\Delta \boldsymbol{\phi}_{i,j}]^\wedge)}& = \exp{([- \boldsymbol{\phi}_i^W]^\wedge)}\exp{([ \boldsymbol{\phi}_j^W]^\wedge)}\\
 \Delta \mathbf{P}_{i,j} &= \exp{([- \boldsymbol{\phi}_i^W]^\wedge)}(\mathbf{P}_j^W-\mathbf{P}_i^W-\mathbf{V}_i^W\delta t_{i,j}-\frac{1}{2}\mathbf{g}\delta t_{i,j}^2) \\
\Delta \mathbf{V}_{i,j} &=\exp{([- \boldsymbol{\phi}_i^W]^\wedge)} \cdot (\mathbf{V}_j^W-\mathbf{V}_i^W-\mathbf{g}\delta t_{i,j}),
\end{split} }
\end{equation}
where the preintegration measurement has the preintegrated noise covariance $\mathbf{\Sigma}_{I}$ (see detailed derivation in \cite{forster2016manifold}).

After bias correction,  $\exp{([\Delta \boldsymbol{\phi}_{i,j}]^\wedge)}$, $\Delta \mathbf{P}_{i,j}$, and $\Delta \mathbf{V}_{i,j}$ update to $\exp([\Delta \widetilde{\boldsymbol{\phi}_{i,j}}]^\wedge)$, $\Delta\widetilde{ \mathbf{P}_{i,j}}$, and $\Delta\widetilde{\mathbf{V}_{i,j}}$. The error term $\mathbf{r}^I$ for optimization based on the preintegrated IMU measurements can be computed as:

\begin{equation}
\label{residual}
\scalemath{0.8}{
\begin{bmatrix}
\mathbf{r}_{\phi}\\
\mathbf{r}_p\\
\mathbf{r}_v\\
\mathbf{r}_{b_a}\\
\mathbf{r}_{b_g}
\end{bmatrix}=\begin{bmatrix} 
\log{[\exp{([-\Delta  \widetilde{\boldsymbol{\phi}_{i,j}}]^\wedge)}\cdot \exp{([- \boldsymbol{\phi}_i^W]^\wedge)} \cdot \exp{([\boldsymbol{\phi}_j^W]^\wedge)}]}^\vee\\
\exp{([- \boldsymbol{\phi}_i^W]^\wedge)} \cdot (\mathbf{P}_j^W-\mathbf{P}_i^W-\mathbf{V}_i^W\delta t_{i,j}-\frac{1}{2}\mathbf{g}\delta t_{i,j}^2)- \Delta  \widetilde{\mathbf{P}_{i,j}}\\
\exp{([-\boldsymbol{\phi}_i^W]^\wedge)} \cdot (\mathbf{V}_j^W-\mathbf{V}_i^W-\mathbf{g}\delta t_{i,j})- \Delta  \widetilde{\mathbf{V}_{i,j}} \\
\mathbf{b}_{j}^a-\mathbf{b}_{i}^a\\
\mathbf{b}_{j}^g-\mathbf{b}_{i}^g
\end{bmatrix}.}
\end{equation}

Then, we compute the jacobian matrix $\mathbf{J}$, which is the derivative of $\mathbf{r}^I$ with respect to vector $[\delta\boldsymbol{\phi}_i^W, \delta\mathbf{P}_i^W,  \delta\mathbf{V}_i^W, \delta\mathbf{b}_{i}^a,  \delta\mathbf{b}_{i}^g, \delta\boldsymbol{\phi}_j^W, \delta\mathbf{P}_j^W, \delta\mathbf{V}_j^W, \delta\mathbf{b}_{j}^a, \delta \mathbf{b}_{j}^g]^T$:

\begin{equation}
\label{jacobian}
\scalemath{0.8}{
\begin{split} 
&\begin{bmatrix} 
\mathbf{J}_{11} &\mathbf{0} &\mathbf{0} &\mathbf{0} &\mathbf{J}_{14} &\mathbf{J}_r^{-1}(r_{\phi}) &\mathbf{0} &\mathbf{0} &\mathbf{0} &\mathbf{0} \\
\mathbf{J}_{21} &-\mathbf{R}_i^T &-\mathbf{R}_i^T\delta t &-\mathbf{J}_{b^a}^p &-\mathbf{J}_{b^g}^p &\mathbf{0} &\mathbf{R}_i^T  &\mathbf{0}  &\mathbf{0}  &\mathbf{0}\\
\mathbf{J}_{31} &0 &-\mathbf{R}_i^T  &-\mathbf{J}_{b^a}^v &-\mathbf{J}_{b^g}^v &\mathbf{0} &\mathbf{0} & \mathbf{R}_i^T  &\mathbf{0}   &\mathbf{0}\\
\mathbf{0} &\mathbf{0} &\mathbf{0} &\mathbf{0} &-\mathbf{I} &\mathbf{0} &\mathbf{0} &\mathbf{0} &\mathbf{I} &\mathbf{0}\\
\mathbf{0} &\mathbf{0} &\mathbf{0} &\mathbf{0} &\mathbf{0} &-\mathbf{I} &\mathbf{0} &\mathbf{0} &\mathbf{0} &\mathbf{I}
\end{bmatrix},
\end{split}}
\end{equation}
where
\begin{equation}
\scalemath{0.8}{
\begin{split}
&\mathbf{J}_{11} =-\mathbf{J}_r^{-1}(r_{\phi})\mathbf{R}_i^T \mathbf{R}_j\\
&\mathbf{J}_{14}= -\mathbf{J}_r^{-1}(r_{\phi})\mathbf{R}_j^T\mathbf{R}_i^T \Delta \widetilde{\mathbf{R}}_{i,j}\mathbf{J}_{b^g}^R\\
&\mathbf{J}_{21}= (\mathbf{R}_i^T (\mathbf{P}_j^W-\mathbf{P}_i^W-\mathbf{V}_i^W\delta t-\frac{1}{2}\mathbf{g}\delta t_{i,j}^2))^\wedge \\
&\mathbf{J}_{31}=(\mathbf{R}_i^T (\mathbf{V}_j^W-\mathbf{V}_i^W-\mathbf{g}\delta t_{i,j}))^\wedge \\
&\mathbf{J}_r^{-1}(r_\phi)=\mathbf{I} + \frac{1}{2}r_\phi^\wedge+(\frac{1}{{\parallel r_\phi \parallel}^2}-\frac{1+\cos{(\parallel r_\phi \parallel)}}{2\parallel r_\phi \parallel \sin{(\parallel r_\phi \parallel)}})(r_\phi^\wedge)^2.
\end{split}}
\end{equation}

\subsection{Feature Extraction}

The feature extraction process is similar to LOAM \cite{zhang2014loam}. We denote the $j^{th}$ LiDAR scan as $\mathcal{P}_j$, and each point as $\mathbf{p}_j^{(m,n)}$. We can evaluate the smoothness of point $\mathbf{p}_j^{(m,n)}$ in local surface $\mathcal{C}_j^{(m,n)}$ by:
\begin{equation}
\label{deqn_smooth}
\scalemath{0.8}{
 \sigma_j^{(m,n)}=\frac{1}{\lvert\mathcal{C}_j^{(m,n)}\rvert\lVert \mathbf{p}_j^{(m,n)} \rVert} \sum_{{\mathbf{p}_j^{(m,l)}} \in {\mathcal{C}_j^{(m,n)}}} {\lVert \mathbf{p}_j^{(m,l)}-\mathbf{p}_j^{(m,n)}\rVert}}.
\end{equation}

The edge feature points set $E_j$ is selected with $\sigma$ larger than threshold $\sigma_{th}$, and the surface feature points set $S_j$ is selected with $\sigma$ smaller than $\sigma_{th}$.

\subsection{Motion Distortion}
 
Similar to FLOAM \cite{wang2021f}, we use the two-stage distortion compensation, which can improve computational efficiency in comparison to iterative motion estimation. 
In the first stage, we compute the relative LiDAR motion between two scans by preintegrating IMU measurements during a LiDAR sweep time $\delta_{t_L}$. Assume that the relative motion estimated from the IMU preintegration and the state at the last scan is $T_j(\delta_{t_L}^{(m,n)})$ for the point $\mathbf{p}_j^{(m,n)}$, the de-skewed current scan $\mathcal{P}_j$ can be written as $\widetilde{\mathcal{P}_j}=\{T_j(\delta_{t_L}^{(m,n)})\mathbf{p}_j^{(m,n)} \arrowvert    \mathbf{p}_j^{(m,n)}\in \mathcal{P}_j\}$.

\subsection{State Estimation With SE(2) Constraints}
\label{State Estimation With SE(2) Constraints}

After motion distortion compensation, we can obtain the undistorted edge features $\tilde{E_j}$ and planar features $\tilde{S_j}$. The aim of motion estimation is to compute the transform matrix $\mathbf{T}$ between the current frame to the global map, which consists of an edge feature map and a planar feature map. Similar to LOAM \cite{zhang2014loam}, we compute the point-to-edge residual and point-to-plane residual by minimizing the distance from the target edge feature point to its corresponding line and the target planar feature point to its corresponding plane. Noticing given an edge feature point and planar feature point in the local LiDAR frame, while the points of $\tilde{E_j}$ and $\tilde{S_j}$ are registered in the global map, we need to transform each local feature point $\mathbf{p}_j$ to the global map by:
\begin{equation}
\label{deqn_l2g}
\hat{\mathbf{p}}_j=\mathbf{T}_j\mathbf{p}_l=\mathbf{R}_j\mathbf{p}_j+\mathbf{P}_j.
\end{equation}
where matrices $\mathbf{R}_j$ and $\mathbf{P}_j$ can be retrieved from $SE(2)$ variables:
\begin{equation}
\label{deqn_ex8}
\mathbf{R}_j=\exp{([0\quad 0\quad \boldsymbol{\phi}_j]^T)} \quad\mathbf{ P}_j=[\mathbf{d}_j^T\quad 0] .
\end{equation}

Then we can get the edge point $\hat{\mathbf{p}}_j\in\tilde{E_j}$. Once we find two nearest points $\mathbf{p}_a^E$ and $\mathbf{p}_b^E$ from the global map, the point-to-edge residual can be computed as:
\begin{equation}
\label{deqn_p2e}
f_E(\hat{\mathbf{p}}_j)=\frac{\lvert(\hat{\mathbf{p}}_j-\mathbf{p}_b^E)\times(\hat{\mathbf{p}}_j-\mathbf{p}_a^E)\rvert}{\lvert\mathbf{p}_a^E-\mathbf{p}_b^E\rvert}.
\end{equation}

Similar to the edge feature points, for a planar feature point $\hat{\mathbf{p}}_j \in \tilde{S_j}$, we search three nearest points $\mathbf{p}_a^S$, $\mathbf{p}_b^S$ and $\mathbf{p}_c^S$. The point-to-plane residual is defined by:
\begin{equation}
\label{deqn_p2p}
f_S(\hat{\mathbf{p}}_j)= \Big\lvert(\hat{\mathbf{p}}_j-\mathbf{p}_a^S)^T\frac{(\mathbf{p}_a^S-\mathbf{p}_b^S)\times(\mathbf{p}_c^S-\mathbf{p}_a^S)}{\lvert(\mathbf{p}_a^S-\mathbf{p}_b^S)\times(\mathbf{p}_c^S-\mathbf{p}_a^S)\rvert}\Big\rvert.
\end{equation}

Then the residual term $\mathbf{r}^L$ for optimization based on LiDAR measurements can be computed as: 
\begin{equation}
\label{deqn_GN}\mathbf{r}^L=\sum_{\hat{\mathbf{p}}_j\in\tilde{E_j}}f_E(\mathbf{T}_j*\mathbf{p}_j)+\sum_{\hat{\mathbf{p}}_j\in\tilde{S_j}}f_S(\mathbf{T}_j*\mathbf{p}_j).
\end{equation}

Although LiDAR poses are parameterized on SE(2) for ground vehicles, the robot's motion in real-world environments is often out-of-SE(2) constraints model due to rough terrains or motion shaking \cite{8793928}.
To solve this problem, we propose a novel method to solve the non-SE(2) perturbations, which directly parameterize the ground vehicles' poses without neglecting the out-of-SE(2) motion perturbations. We incorporate the out-of-SE(2) perturbations as noise into (\ref{deqn_l2g}). Assume the vertical translation perturbation is $\eta_z \sim \mathcal{N}(0,\sigma_z^2)$, the rotation perturbations are $\eta_{\theta_{xy}} \sim \mathcal{N}(\textbf{\textit{0}}_{2\times1},\mathbf{\Sigma}_{\theta_{xy}})$, and the perturbed pose is like:
\begin{equation}
\label{deqn_ex9}
\mathbf{R}_j\leftarrow \exp{(\underbrace{[\boldsymbol{\eta}_{\theta_{xy}}^T \quad 0]^T}_{\boldsymbol{\eta}_\theta})}\mathbf{R}_j \quad \mathbf{P}_j\leftarrow \mathbf{P}_j+\underbrace{[0\quad 0 \quad \eta_z]^T}_{\boldsymbol{\eta}_z}.
\end{equation}

The Jacobian of $f_E$ w.r.t $({\eta_\theta},\eta_z)$ can be estimated by applying the right perturbation model:


\begin{equation}
\label{deqn_ex10}
\scalemath{0.85}{
\begin{split}
\mathbf{J}_{\eta_\theta}^E &=\frac{\partial f_E}{\partial \mathbf{T}\mathbf{p}_j}\frac{\partial \mathbf{T}\mathbf{p}_j}{\partial \delta\boldsymbol{\eta}_\theta}\\
&=\frac{\partial f_E}{\partial \mathbf{T}\mathbf{p}_j}\lim_{\delta\boldsymbol{\eta}_\theta \to 0}\frac{\mathbf{T}\exp{(\delta\boldsymbol{\eta}_\theta^\wedge)}\mathbf{p}_j-\mathbf{T}\mathbf{p}_j}{\delta\boldsymbol{\eta}_\theta}\\
&=\frac{((\hat{\mathbf{p}}_j-\mathbf{p}_b^E)\times(\hat{\mathbf{p}}_j-\mathbf{p}_a^E))^T}{\lvert(\hat{\mathbf{p}}_j-\mathbf{p}_b^E)\times(\hat{\mathbf{p}}_j-\mathbf{p}_a^E)\rvert} \frac{(\mathbf{p}_a^E-\mathbf{p}_b^E)^\wedge}{\lvert\mathbf{p}_a^E-\mathbf{p}_b^E\rvert}(-\mathbf{R}(\mathbf{p}_j)^\wedge),
\end{split} }
\end{equation}

\begin{equation}
\label{deqn_ex11}
\scalemath{0.9}{
\begin{split}
\mathbf{J}_{\eta_z}^E &=\frac{\partial f_E}{\partial \mathbf{T}\mathbf{p}_j}\frac{\partial \mathbf{T}\mathbf{p}_j}{\partial \delta\boldsymbol{\eta}_z}\\
&=\frac{((\hat{\mathbf{p}}_j-\mathbf{p}_b^E)\times(\hat{\mathbf{p}}_j-\mathbf{p}_a^E))^T}{\lvert(\hat{\mathbf{p}}_j-\mathbf{p}_b^E)\times(\hat{\mathbf{p}}_j-\mathbf{p}_a^E)\rvert} \frac{(\mathbf{p}_a^E-\mathbf{p}_b^E)^\wedge}{\lvert\mathbf{p}_a^E-\mathbf{p}_b^E\rvert}.
\end{split} }
\end{equation}

Similarly, for surface residual $f_S$ we can derive Jacobian w.r.t $({\eta_\theta},\eta_z)$ :

\begin{equation}
\label{deqn_ex12}
\mathbf{J}_{\eta_\theta}^S= \frac{((\mathbf{p}_a^S-\mathbf{p}_b^S)\times(\mathbf{p}_c^S-\mathbf{p}_a^S))^T}{\lvert(\mathbf{p}_a^S-\mathbf{p}_b^S)\times(\mathbf{p}_c^S-\mathbf{p}_a^S)\rvert} (-\mathbf{R}(\mathbf{p}_j)^\wedge).
\end{equation}

\begin{equation}
\label{deqn_ex13}
\mathbf{J}_{\eta_z}^S=\frac{((\mathbf{p}_a^S-\mathbf{p}_b^S)\times(\mathbf{p}_c^S-\mathbf{p}_a^S))^T}{\lvert(\mathbf{p}_a^S-\mathbf{p}_b^S)\times(\mathbf{p}_c^S-\mathbf{p}_a^S)\rvert}.
\end{equation}

Assume LiDAR sensor noise is $\eta_k \sim \mathcal{N}(0,\sigma_k^2)$. Then, we formulate a synthetic zero-mean noise for both edge and surface features $\delta\boldsymbol{\eta}=\mathbf{J}_{\eta_\theta}\boldsymbol{\eta}_\theta + \mathbf{J}_{\eta_z}\boldsymbol{\eta}_z + \boldsymbol{\eta}_k$, where $\boldsymbol{\eta}_\theta$, $\boldsymbol{\eta}_z$, $\boldsymbol{\eta}_k$ are independent with each other. Therefore, the covariance matrix can be approximately expressed as:
\begin{equation}
\label{deqn_ex14}
\boldsymbol{\Sigma_L}=\mathbf{J}_{\eta_\theta}\boldsymbol{\wedge}_{12}\boldsymbol{\Sigma}_{\theta_{xy}}\mathbf{J}_{\eta_\theta}^T+\sigma_z^2\mathbf{J}_{\eta_z}\mathbf{e}_3\mathbf{e}_3^T\mathbf{J}_{\eta_z}^T+\sigma_k^2,
\end{equation}
where $\mathbf{e}_i$ denotes the $i^{th}$ column of the matrix $\mathbf{I}_{3\times3}$, and $\boldsymbol{\wedge}_{12}=[\mathbf{e}_1\quad \mathbf{e}_2]$.

We can now construct the residual error term of the SE(2)-XYZ constraints, the Jacobian of edge and surface residuals w.r.t $(\boldsymbol{\phi}, \mathbf{d})$ can be estimated by applying the right perturbation model with $\delta\boldsymbol{\xi}\in\mathfrak{se}(2)$.

\begin{equation}
\label{deqn_ex15}
\mathbf{J}_p=\frac{\partial \mathbf{T}\mathbf{p}_j}{\partial\delta\boldsymbol{\xi}}=\left[\mathbf{I}_{3\times3}\boldsymbol{\wedge}_{12} \quad \textbf{\textit{0}}_{1\times3} \quad -\mathbf{R}(\mathbf{p}_j)^\wedge \mathbf{e}_3\right].
\end{equation} 

The Jacobian matrix of edge residual is calculated as:
\begin{equation}
\label{deqn_ex16}
\mathbf{J}_E=\frac{\partial f_E}{\partial \mathbf{T}\mathbf{p}_j}\frac{\partial \mathbf{T}\mathbf{p}_j}{\partial\delta\boldsymbol{\xi}}=\frac{((\hat{\mathbf{p}}_j-\mathbf{p}_b^E)\times(\hat{\mathbf{p}}_j-\mathbf{p}_a^E))^T}{\lvert(\hat{\mathbf{p}}_j-\mathbf{p}_b^E)\times(\hat{\mathbf{p}}_j-\mathbf{p}_a^E)\rvert} \mathbf{J}_p.
\end{equation}

Similarly, we can derive:
\begin{equation}
\label{deqn_ex17}
\mathbf{J}_S= \frac{((\mathbf{p}_a^S-\mathbf{p}_b^S)\times(\mathbf{p}_c^S-\mathbf{p}_a^S))^T}{\lvert(\mathbf{p}_a^S-\mathbf{p}_b^S)\times(\mathbf{p}_c^S-\mathbf{p}_a^S)\rvert} \mathbf{J}_p.
\end{equation}


The optimal pose estimation can be derived by solving the non-linear equation through the Gauss-Newtons method:
\begin{equation}
    \label{least-squares}
    \widehat{\mathcal{X}}=\mathop{\arg\min}_{\mathcal{X}} \Big\{\sum \| \mathbf{r}^I\|^2_{\mathbf{\Sigma}_I} +  \sum \| \mathbf{r}^L\|^2_{\mathbf{\Sigma}_L}\Big\} .
\end{equation}

The  complete iterative pose estimation algorithm is summarized in Algorithm \ref{fig: algorrithm1}.
\begin{algorithm}[t]
\SetAlgoLined
\SetKwInOut{Input}{Input}
\SetKwInOut{Output}{Output}
 \Input{IMU measurements $\hat{\mathbf{w}}_t$ and $\hat{\mathbf{a}}_t$, current scan $\mathcal{P}_{j}$, state $\mathcal{X}_{i}$}
 \Output{Current state $\mathcal{X}_{j}$}
\Begin{
    \lIf{Not Initialized}{$\mathcal{X}_{0} \leftarrow \textbf{0}$, $\mathbf{\Sigma}_{I}\leftarrow \textbf{0}_{15\times15}$}
    Calculate the robot motion $\exp{([\Delta \boldsymbol{\phi}_{i,j}]^\wedge)}$, $\Delta \mathbf{P}_{i,j}$, and $\Delta \mathbf{V}_{i,j}$ from time $i$ to $j$ by IMU preintegration\;
    Calculate the local smoothness and extract edge features and planar features \;
    Compute de-skewed LiDAR scan $\widetilde{\mathcal{P}_j}$ \;
    \For{Pose optimization is not converged}{
       Compute the IMU residual $\mathbf{r}^I$ and jacobian $\mathbf{J}$ with covariance matrix $\mathbf{\Sigma}_{I}$\;
        \For{each point $\mathbf{p}_{j} \in \mathcal{P}_{j}$}{
            Transform to map coordinate 
            $\hat{\mathbf{p}}_j \leftarrow \textbf{T}_{j}\textbf{p}_j$\;
            
            \If{$\mathbf{p}_{j}$ is an edge feature}{
                Compute edge residual $f_{E}$ and Jacobian matrix $J_{E}$ with $\boldsymbol{\Sigma_L}$ \;
            }
            \If{$\mathbf{p}_{j}$ is a planar feature}{
                Compute planar residual $f_{S}$ and Jacobian matrix $J_{S}$ with $\boldsymbol{\Sigma_L}$\;
            }
        }
        \If{nonlinear optimization converges}{
            Compute $\mathcal{X}_{j}$\;
        }
    }    
    Update bias $\mathbf{b^a}$ and $\mathbf{b^g}$, covariance matrix $\mathbf{\Sigma}_{I}$ and update current scan into local map\;
} 
\caption{Pose estimation for SE2LIO}
\label{fig: algorrithm1}
\end{algorithm}

\begin{figure}[t]
     \centering
     \begin{subfigure}[]{0.49\linewidth}
         \centering
         \includegraphics[width=\linewidth]{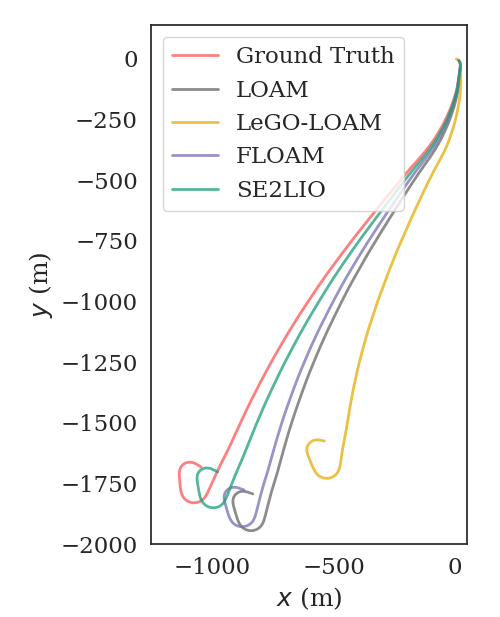}
         \caption{KITTI Sequence 01}
         \label{fig:s01}
     \end{subfigure}
     \begin{subfigure}[]{0.49\linewidth}
         \centering
         \includegraphics[width=\linewidth]{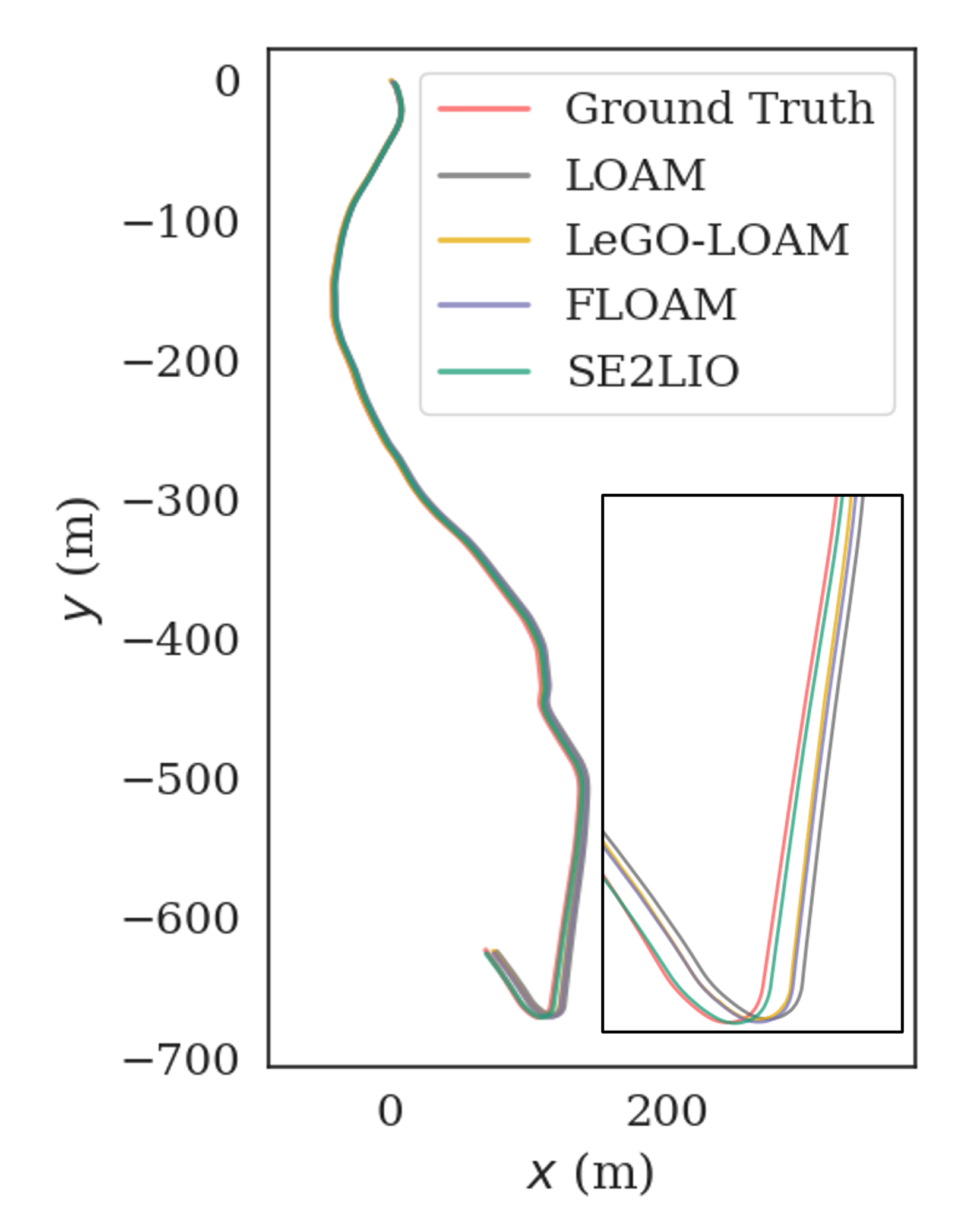}
         \caption{KITTI Sequence 10}
         \label{fig:s10}
     \end{subfigure}
     \hfill
     \begin{subfigure}[]{1.0\linewidth}
         \centering
         \includegraphics[width=\linewidth]{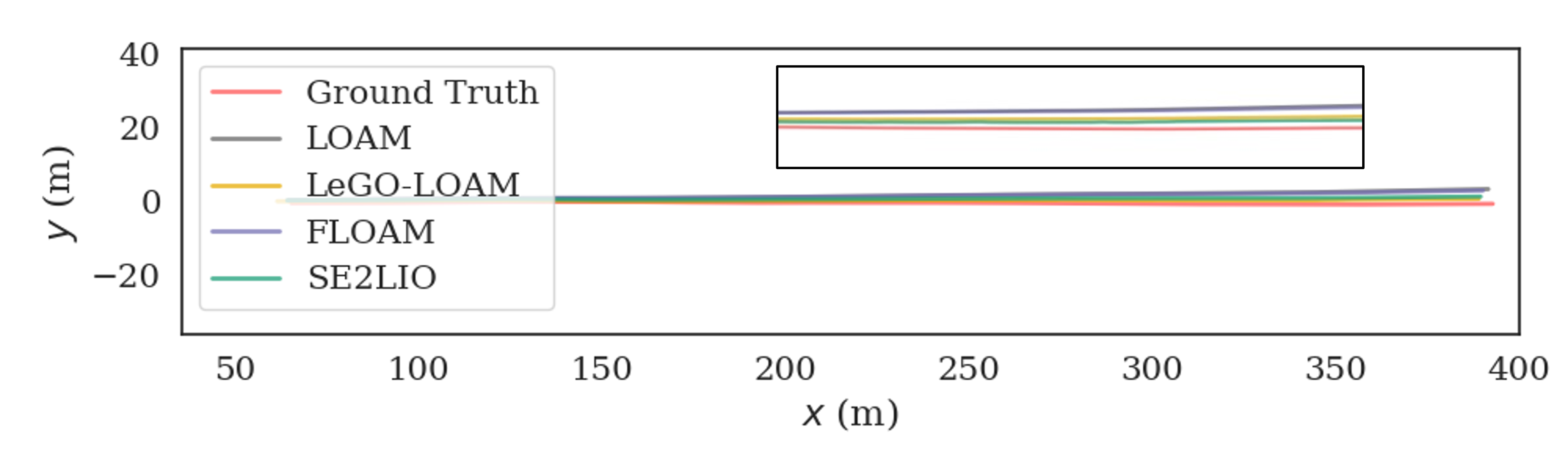}
         \caption{KITTI Sequence 04}
         \label{fig:s04}
     \end{subfigure}
        \caption{
        The estimated trajectories for KITTI sequence 01, 04, and 10.}
        \label{fig_traj}
\end{figure}

\begin{figure}[t]
     \centering
     \begin{subfigure}{1.0\linewidth}
         \centering
         \includegraphics[width=\linewidth]{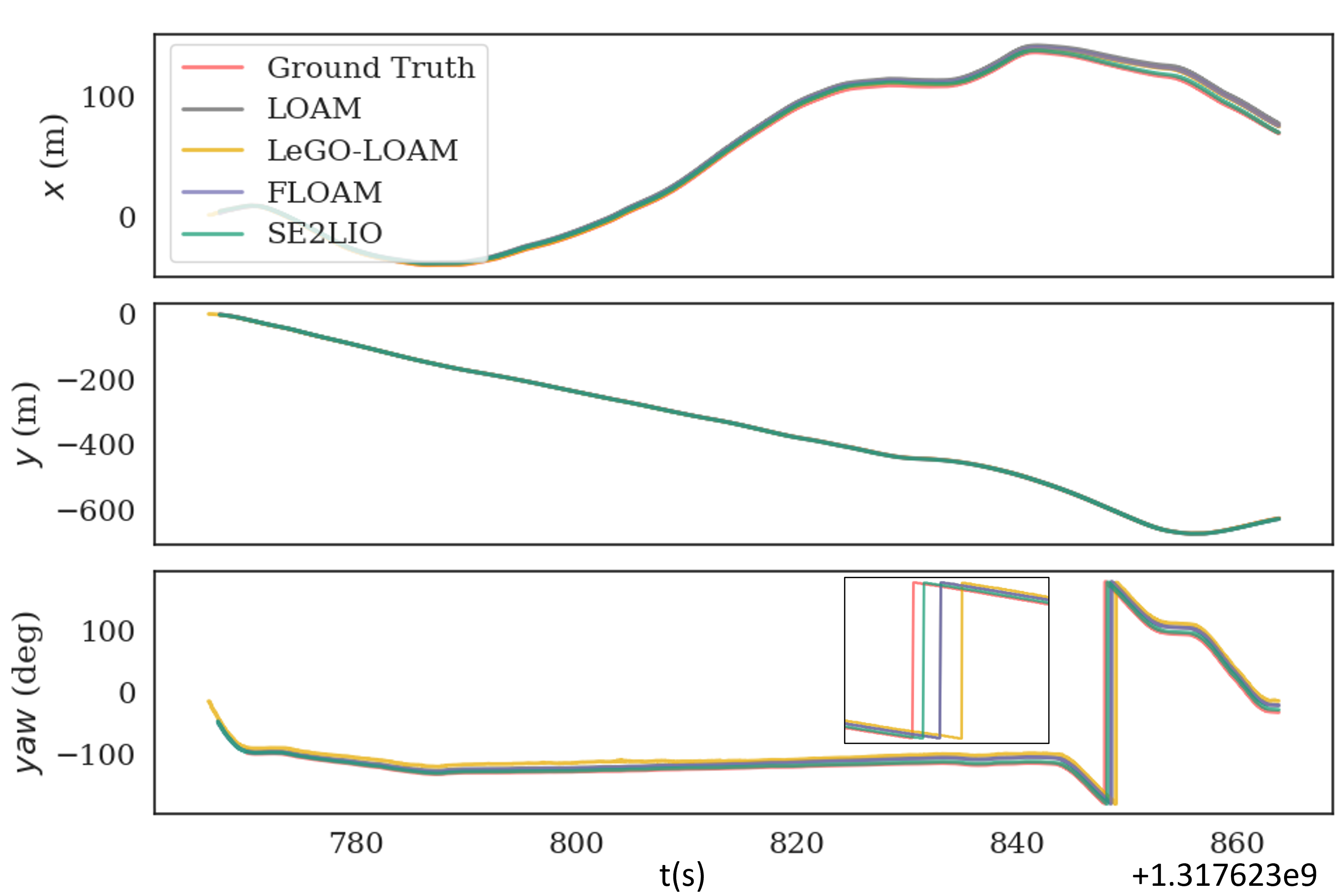}
         \caption{KITTI Sequence 01}
         \label{fig:s01xyr}
     \end{subfigure}
     \hfill
     \begin{subfigure}{1.0\linewidth}
         \centering
         \includegraphics[width=\linewidth]{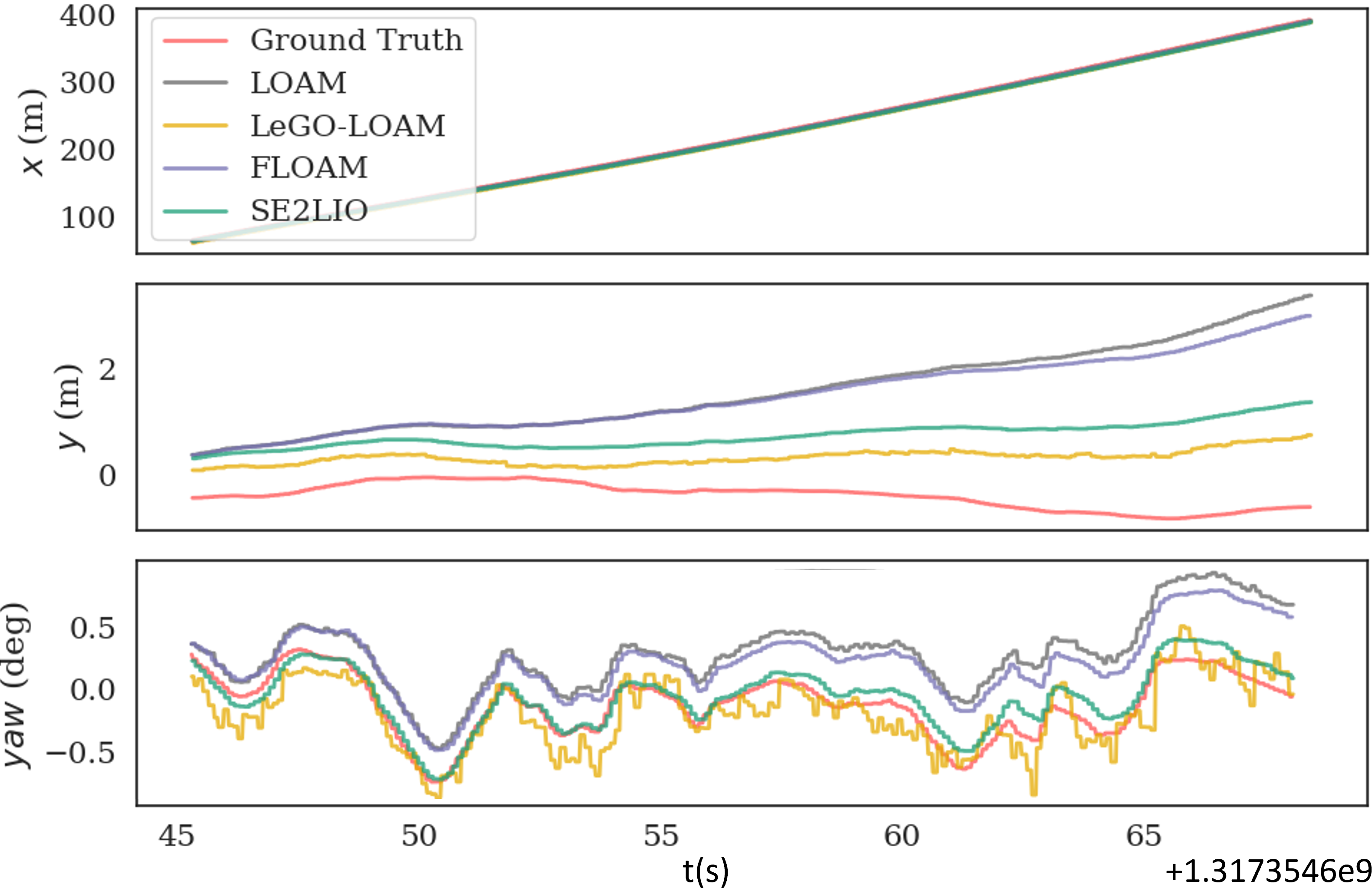}
         \caption{KITTI Sequence 04}
         \label{fig:s04xyr}
     \end{subfigure}
     \hfill
     \begin{subfigure}{1.0\linewidth}
         \centering
         \includegraphics[width=\linewidth]{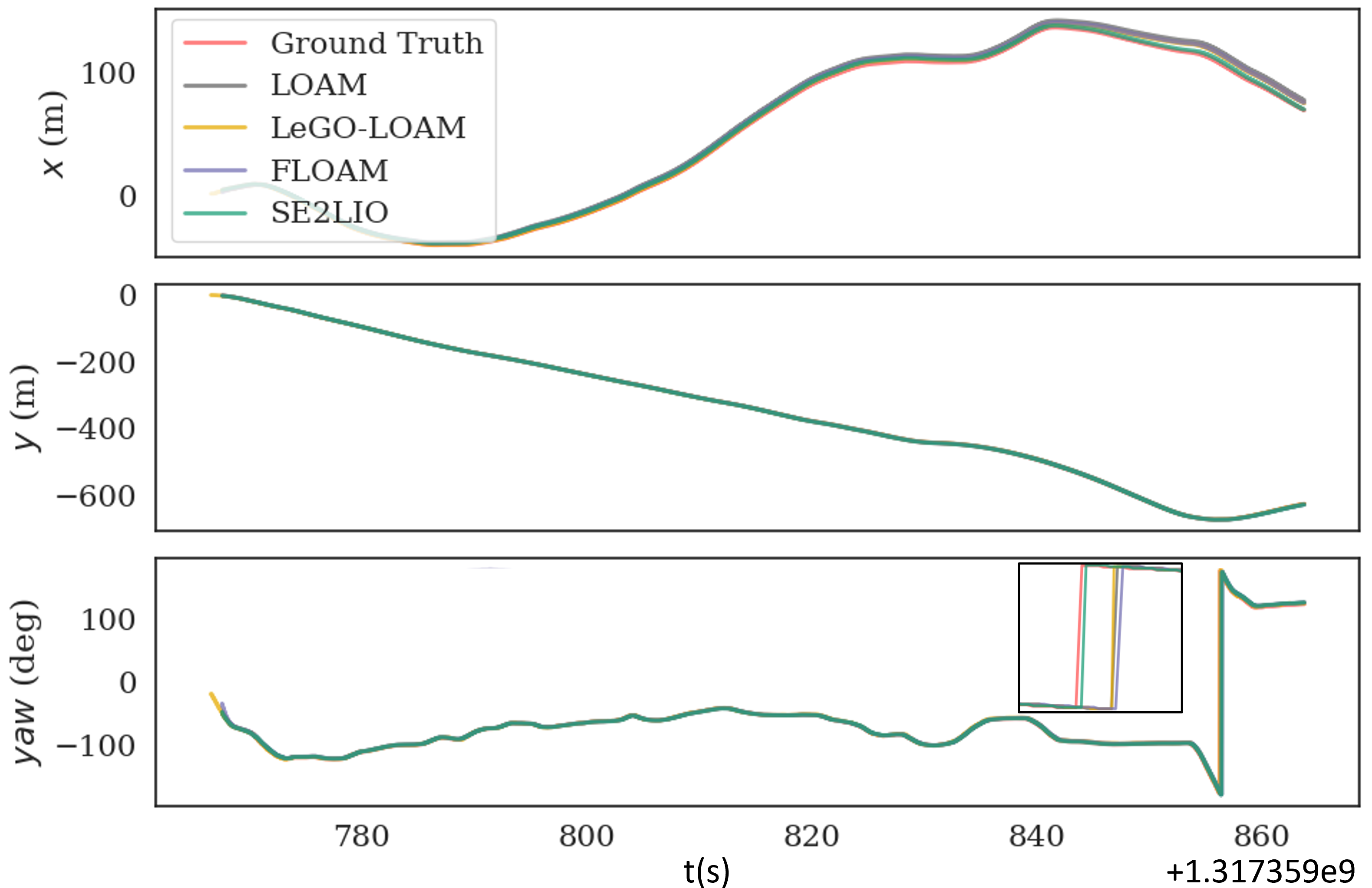}
         \caption{KITTI Sequence 10}
         \label{fig:s10xyr}
     \end{subfigure}
        \caption{KITTI dataset evaluation: the translation on x,y direction and rotation around z-axis with respectively to KITTI sequence 01, 04, and 10.}
        \label{fig_trans_rot}
\end{figure}

\subsection{Mapping and Distortion Compensation Update}
\label{Mapping and Distortion Compensation Update}
After motion estimation, the local edge map and the local plane maps are generated, which are stored in 3D KD-trees and used to build the global map. In the second stage of distortion compensation, the distortion is re-computed based on the optimization result $\mathbf{T}_j^*$ from the last step by:

\begin{equation}
\label{deqn_ex18}
\begin{aligned}
\Delta\boldsymbol{\xi}^* &= \log{(\mathbf{T}_{i}^{-1}\mathbf{T}_j^*)} \\
 \mathcal{P}_j^* &= \{\exp{(\frac{N-n}{N} \Delta\boldsymbol{\xi}^*)}\mathbf{p}_j^{(m,n)} \arrowvert \mathbf{p}_j^{(m,n)}\in  \mathcal{P}_j\} .
\end{aligned}
\end{equation}

\section{EXPERIMENT EVALUATION}\label{section:EXPERIMENT EVALUATION} 
To validate our proposed method, we evaluate the algorithm in both outdoor environments and indoor environments. Firstly, we evaluate the performance of the proposed method with SE(2) constraints on the KITTI dataset \cite{geiger2012we}. We also provide experiments in the indoor environment. Then, we evaluate the SE(2) constraints with tightly-coupled IMU constraints in the gravel car park and large-scale urban environments. All the experiment results are compared with SOTA methods, and all methods are executed using ROS Melodic \cite{quigley2009ros} in Ubuntu 18.04. 
\textcolor{blue}{}

 
\subsection{KITTI Dataset}
The KITTI dataset is collected from a passenger vehicle equipped with stereo cameras, a Velodyne HDL-64 LiDAR, and a high-accuracy GPS for ground truth purposes. The LiDAR data is logged at 10Hz with various scenarios such as city, highway, campus, etc. Note that the IMU frequency is only provided at 10Hz on the KITTI dataset, only SE(2) constraint is applied. We evaluate the proposed method on the KITTI sequence 00-11 with GPS data as ground truth. Some of the experimental results, in which sequences 01, 04, and 10 are selected for demonstration.

Fig. \ref{fig_traj} demonstrates the trajectory results of KITTI dataset sequences 01, 04, and 10, respectively, the results are estimated by our proposed system with SE(2) constraints comparison with LOAM, LeGO-LOAM, FLOAM, and all methods are tested on a laptop with an Intel i7 2.9GHz processor. Since the mapping process is the same as FLOAM, only some of the mapping results are shown. The corresponding orientation and translation error computed w.r.t the ground truth and the computational cost are shown in Table \ref{kitti}. The computational cost is measured by the average runtime of each frame. To evaluate the localization accuracy, we use the Root Mean Square Error (RMSE) of the Absolute Translational Error (ATE) and Absolute Rotational Error (ARE) \cite{sturm2012benchmark}. The ATE and ARE are based on the absolute relative pose between the estimated pose $\mathbf{T}_{est, i}$ and the ground truth $\mathbf{T}_{gt, i}$:
\begin{equation}
\label{deqn_error}
\begin{aligned}
ATE_i&=\lvert trans(\mathbf{T}_{gt,i}^{-1}\mathbf{T}_{est,i})\rvert \\
ARE_i&=\lvert\angle{(\mathbf{T}_{gt,i}^{-1}\mathbf{T}_{est,i})}\rvert,
\end{aligned}
\end{equation}
where $trans(.)$ and $\angle{(.)}$ represent the translation part and rotation angle of $\mathbf{T}$.

\begin{table}[t]
\caption{State Estimation Error on KITTI Dataset }
\begin{center}
\begin{tabular}{lcccc}
\multicolumn{1}{l|}{}             & LOAM & LeGO-LOAM & FLOAM & SE2LIO \\ \hline
Sequence 01           &          &       &        &                        \\ \hline
\multicolumn{1}{l|}{trans.(m)}  &168.36  &393.76  &138.11  &\textbf{51.67}                        \\
\multicolumn{1}{l|}{rot.(deg)} & 12.46   &16.15  &7.83        &\textbf{2.30}                                        \\ 
\multicolumn{1}{l|}{time(ms/frame)} & 100.45   & 81.43   & 70.30        & \textbf{57.73}                             \\ \hline
Sequence 04                       &       &        &      &                   \\ \hline
\multicolumn{1}{l|}{trans.(m)} &    2.19  &4.28  &2.09     & \textbf{1.21}                        \\
\multicolumn{1}{l|}{rot.(deg)} &  0.42  &0.15   &0.36        &\textbf{0.11}                                  \\ 
\multicolumn{1}{l|}{time(ms/frame)} & 86.36   &67.28     & 78.67      & \textbf{58.32}                               \\ \hline
Sequence 10                       &       &        &    &                    \\ \hline
\multicolumn{1}{l|}{trans.(m)} & 5.70 &3.20     &4.03        &\textbf{1.92}                        \\
\multicolumn{1}{l|}{rot.(deg)} &1.45   &1.09    &1.53        &\textbf{1.08}                                            \\ 
\multicolumn{1}{l|}{time(ms/frame)} & 92.71  & 59.20    & 60.52  &  \textbf{47.31}                                   \\ \hline
\end{tabular}
\end{center}
\label{kitti}
\end{table}

\begin{table}[t]
\caption{State Estimation Error on Indoor Dataset}
\begin{adjustbox}{max width=1.0\textwidth,center}
\begin{tabular}{lcccccc}
\multicolumn{1}{l|}{}               & \multirow{2}{*}{\begin{tabular}[c]{@{}c@{}}Wheel\\ Odom.\end{tabular}} & \multirow{2}{*}{LOAM} & \multirow{2}{*}{\begin{tabular}[c]{@{}c@{}}LEGO\\ -LOAM\end{tabular}} & \multirow{2}{*}{\begin{tabular}[c]{@{}c@{}}F-\\ LOAM\end{tabular}} & \multirow{2}{*}{\begin{tabular}[c]{@{}c@{}}SE2\\ -LIO$^{*}$\end{tabular}}&  \multirow{2}{*}{\begin{tabular}[c]{@{}c@{}}SE2\\ -LIO\end{tabular}} \\
\multicolumn{1}{l|}{}               &                                                                        &                       &                                                                       &                        &                         \\ \hline
Warehouse                           & \multicolumn{1}{l}{}                                                   & \multicolumn{1}{l}{}  & \multicolumn{1}{l}{}                                                  & \multicolumn{1}{l}{}   & \multicolumn{1}{l}{}    \\ \hline
\multicolumn{1}{l|}{trans.(m)}      & 0.133                                                                  & 0.169                 & 0.201                                                                 & 0.162          &0.121       & \textbf{0.084}          \\
\multicolumn{1}{l|}{rot.(deg)}      & 1.417                                                                  & 9.688                 & 9.24                                                                  & 9.523           &7.721       & \textbf{0.867}          \\
\multicolumn{1}{l|}{time}    & \multirow{2}{*}{-} & \multirow{2}{*}{87.91}                                                & \multirow{2}{*}{\textbf{55.13} }                                             & \multirow{2}{*}{75.71}  &\multirow{2}{*}{62.12} & \multirow{2}{*}{64.32 }     \\
\multicolumn{1}{l|}{(ms/frame)}  &                         &                                                                       &                                                                     &                        &                            \\ \hline
\end{tabular}
\end{adjustbox}
\label{warehouse}
\end{table}

As can be seen from Table \ref{kitti}, the SE2LIO system shows higher accuracy in translation and rotation. For KITTI sequence 01,  the LeGO-LOAM results in a subpar localization performance because the assumption of the sensor model used in LeGO-LOAM is not suitable for KITTI datasets, as well as, extraction of the ground point is noisy and the quality of the ground points in sequence 01 is poor. Compared to the LOAM and FLOAM, it can be seen that the RMSE of ATE by our method is reduced by  $69.3 \%$ and $62.6 \%$. The RMSE of ARE is reduced by $81.5 \%$ and $70.6 \%$. For sequence 04, compared to LOAM, LeGO-LOAM, and FLOAM, the RMSE of ATE is reduced by $44.7 \%$, $71.7 \%$, and $42.1 \%$, respectively. The RMSE of ARE is reduced by $73.8 \%$, $26.7 \%$, and $69.4 \%$. For sequence 10, the RMSE of ATE is reduced by $66.3 \%$, $40.0 \%$, and $52.4 \%$. The RMSE of ARE is reduced by $25.5 \%$, $9.2 \%$, and $29.4 \%$. These results indicate the importance of introducing the planar motion constraints into the state estimation problem when the vehicle moves on a planar surface. For the computational cost, the processing time of our method is less than other methods because of the two-stage distortion compensation \cite{wang2021f} and faster converge rate for SE(2) poses.

\begin{figure}[!tp]
  \vspace*{-\topmargin}\vspace*{-\headsep}\vspace*{-\headheight}\vspace*{-\topskip}
\centering
\includegraphics[width=0.9\linewidth]{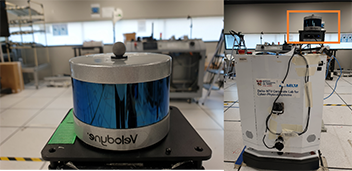}
\caption{AGV platform used for experiment}
\label{fig:agv}
\end{figure}

\begin{figure}[!tp]
\centering
\includegraphics[width=0.99\linewidth]{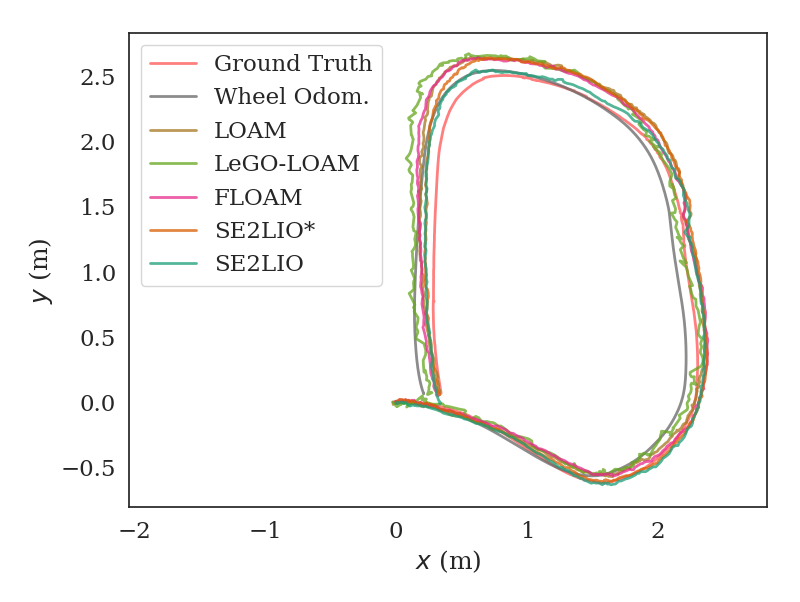}
\caption{
Estimated trajectories for indoor dataset}
\label{fig:lab}
\end{figure}

\subsection{Indoor Dataset}
In the indoor test, we implement the proposed system on an actual AGV equipped with a Velodyne VLP-16 sensor, and an Intel NUC mini computer with an i5-10210U processor for smart manufacturing applications. The robot platform for testing is shown in Fig. \ref{fig:agv}. The brackets used to mount LiDAR are unstable, which causes a lot of non-SE(2) perturbations, while the AGV is moving. Furthermore, the VICON system is used as ground truth to evaluate AGV's localization accuracy. The robot is remotely controlled to move in the testing area. The trajectories results are shown in Fig. \ref{fig:lab}. The estimated translation and rotation errors are shown in Table \ref{warehouse}. Our method reduces the RMSE of ATE by $36.8 \%$, $50.3 \%$, $58.2 \%$, and $48.1 \%$ compared to Wheel Odometry, LOAM, LeGO-LOAM, and FLOAM, respectively. Additionally, our method reduces the RMSE of absolute rotational error ARE by $38.8 \%$, $91.1 \%$, $90.6 \%$, and $90.8 \%$. Moreover, we conduct an experiment SE2LIO$^*$ by only setting perturbation to zero compared to our method SE2LIO with perturbation covariance $10^{-3} rad^2$. SE2LIO reduces the RMSE of ATE and ARE by $30.6 \%$, and $88.8 \%$, respectively, which demonstrates the importance of introducing out-of-SE(2) perturbations.

\subsection{Outdoor Dataset}

This test is designed to show the benefits of introducing the IMU sensor. To evaluate the proposed tightly coupled LiDAR-IMU fusion method, we mount an LPMS-NAV3 IMU (400 Hz) on the AGV and operate it in a gravel car park and a city street. We also collect the GPS measurement as ground truth. For comparison, we run other SOTA methods: LOAM, LeGO-LOAM, FLOAM, and LIO-SAM. Here, LOAM and LeGO-LOAM incorporate a loosely coupled fusion of IMU to de-skew the LiDAR scan.

\begin{figure}[tp]
    \centering
    \begin{subfigure}[]{1.0\linewidth}
           \centering
           \includegraphics[width=0.99\linewidth]{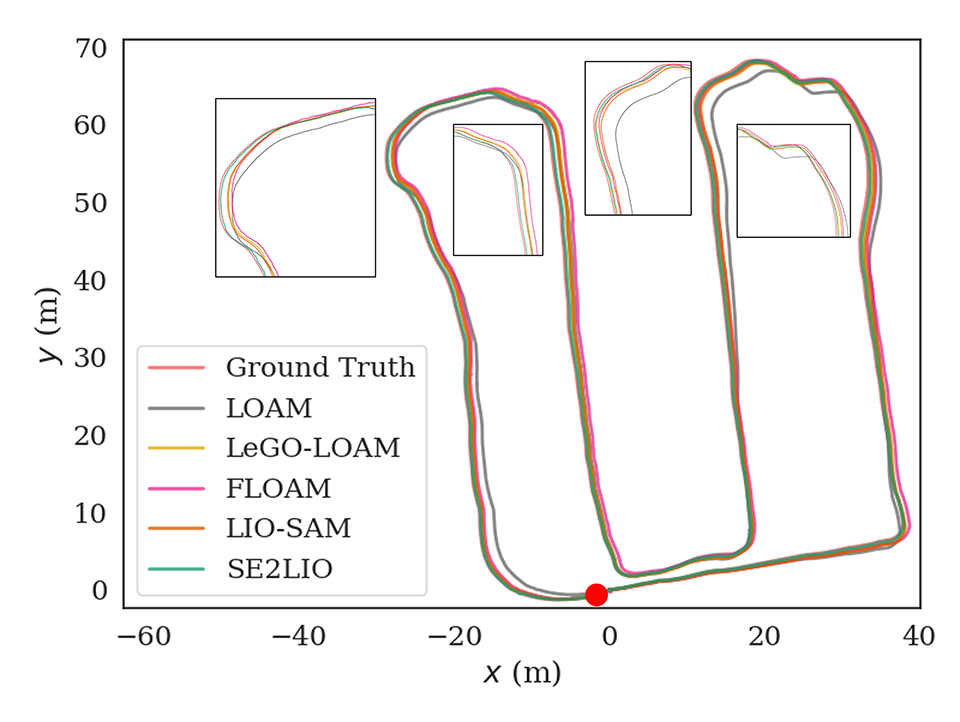}
            \caption{Trajectories from different methods}
            \label{fig:carpark_traj}
    \end{subfigure}
    \begin{subfigure}[]{1.0\linewidth}
            \centering
            \includegraphics[width=0.75\linewidth]{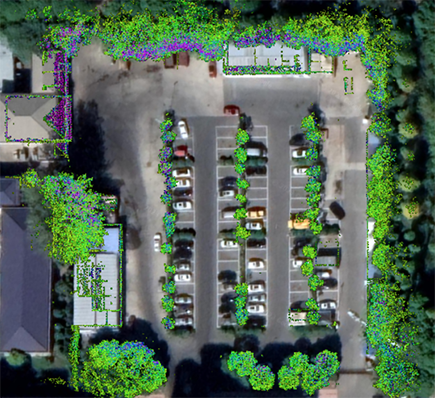}
            \caption{Mapping aligned with Google Earth}
            \label{fig:carpark_mp}
    \end{subfigure}
    \caption{Results of various methods using the car park dataset. The red dot indicates the start and end location. The trajectory direction is counterclockwise.}
    \label{fig: car park}
\end{figure}

\begin{figure}[tp]
    \centering
    \begin{subfigure}[]{1.0\linewidth}
           \centering
           \includegraphics[width=0.99\linewidth]{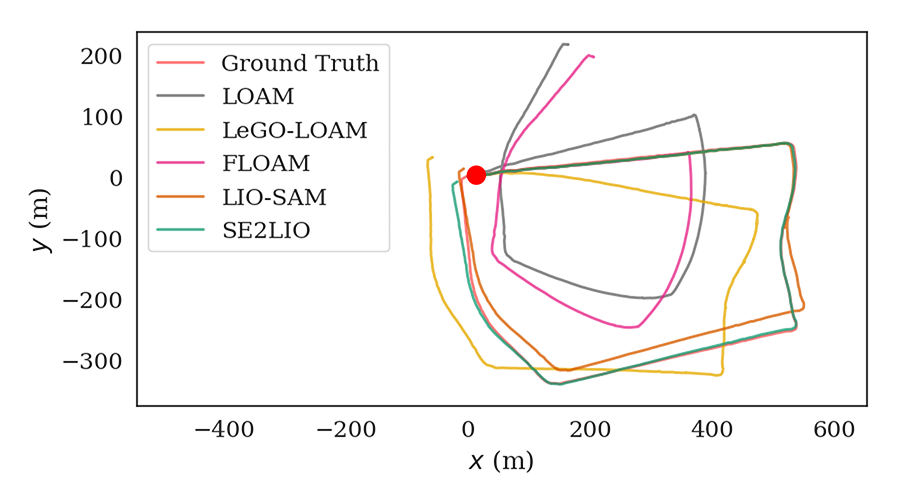}
            \caption{Trajectories from different methods}
            \label{fig:outdoor_traj}
    \end{subfigure}
    \begin{subfigure}[]{1.0\linewidth}
            \centering
            \includegraphics[width=0.75\linewidth]{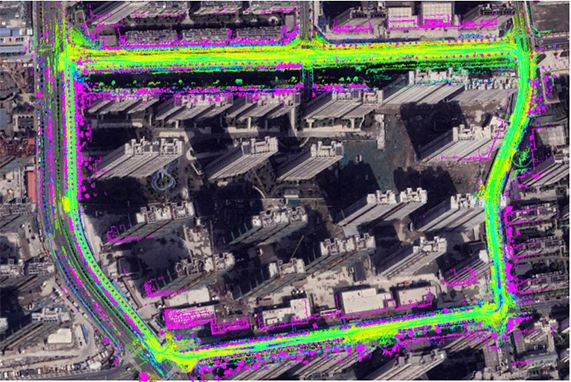}
            \caption{Mapping aligned with Google Earth}
            \label{fig:outdoor_map}
    \end{subfigure}
    \caption{Results of various methods using the city dataset. The red dot indicates the start and end location. The trajectory direction is clockwise.}
    \label{fig: city}
\end{figure}

\begin{table}[h]
\caption{State Estimation Error on Outdoor Datasets}
\begin{tabular}{lccccc}
\multicolumn{1}{l|}{}               & \multirow{2}{*}{LOAM} & \multirow{2}{*}{\begin{tabular}[c]{@{}c@{}}LEGO\\ -LOAM\end{tabular}} & \multirow{2}{*}{FLOAM} & \multirow{2}{*}{\begin{tabular}[c]{@{}c@{}}LIO\\ -SAM\end{tabular}} & \multirow{2}{*}{SE2LIO} \\
\multicolumn{1}{l|}{}               &                       &                                                                       &                        &                                                                     &                      \\ \hline
Car park                           & \multicolumn{1}{l}{}  & \multicolumn{1}{l}{}                                                  & \multicolumn{1}{l}{}   & \multicolumn{1}{l}{}                                                & \multicolumn{1}{l}{}    \\ \hline
\multicolumn{1}{l|}{trans.(m)}      & 1.53                  & 0.63                                                                & 0.97                   &0.70                                                              & \textbf{0.47}          \\
\multicolumn{1}{l|}{rot.(deg)}      & 4.70                  & 3.95                                                                  & 3.96                   & 4.24                                                                & \textbf{1.88}           \\
\multicolumn{1}{l|}{time(ms/frame)} &79.71                 & \textbf{37.42}                                                       & 43.58                 & 51.00                                                               & 53.7                   \\ \hline
City                                & \multicolumn{1}{l}{}  & \multicolumn{1}{l}{}                                                  & \multicolumn{1}{l}{}   & \multicolumn{1}{l}{}                                                & \multicolumn{1}{l}{}    \\ \hline
\multicolumn{1}{l|}{trans.(m)}      & 167.27                & 96.99                                                                 & 150.66                 & 21.12                                                               & \textbf{4.74}           \\
\multicolumn{1}{l|}{rot.(deg)}      & 31.63                 & 16.31                                                                 & 28.83                  & 5.62                                                                & \textbf{2.84}           \\
\multicolumn{1}{l|}{time(ms/frame)} & 116.23                & \textbf{67.2}                                                         & 67.91                  & 85.05                                                               & 88.34                   \\ \hline
\end{tabular}
\label{tab:outdoor}
\end{table}

\begin{table}[!h]
 \centering
\caption{ Ablation study of localization accuracy}
\begin{tabular}{l|ccc}
          & SE2LO & SE2LIO$^\sharp$ & SE2LIO        \\ \hline
trans.(m) & 0.66  & 0.59    & \textbf{0.47} \\
rot.(deg) & 4.27  & 3.88    & \textbf{1.88} \\ \hline
\end{tabular}
\label{tab: abaltion}
\end{table}

\subsubsection{Comparison with SOTA methods}

For the gravel car park dataset, the trajectories and mapping results are shown in Fig. \ref{fig: car park}. The RMSE error w.r.t GPS is shown in Table \ref{tab:outdoor}. The RMSE of ATE by $69.3 \%$, $25.4 \%$, $51.5 \%$, and $32.9 \%$ compared to LOAM, LeGO-LOAM, FLOAM, and LIO-SAM, respectively. Additionally, our method reduces the RMSE of absolute rotational error ARE by $60.0 \%$, $52.4 \%$, $52.5 \%$, and $55.7 \%$. 

For city dataset, the trajectories and mapping results are shown in Fig. \ref{fig: city}. From Table \ref{tab:outdoor}, it can be seen that LOAM and FLOAM provide much less accurate estimation results than our proposed system, further illustrating the influence of ground constraints.
Compared to LeGO-LOAM, the RMSE of ATE and ARE by our proposed LiDAR-inertial odometry system is reduced by $95.1 \%$ and $82.6 \%$ respectively. Moreover, SE2LIO reduces the RMSE of ATE and ARE by $77.6 \% $ and $49.5 \%$ in comparison to LIO-SAM. 
Moreover, the average running time
for each frame is 88 ms which is enough to achieve real-time performance for many robotic applications.

\subsubsection{Ablation study}

To further evaluate the performance of the proposed method, we compare the results of different approaches on the car park dataset as shown in Table \ref{tab: abaltion}. Compared to SE2LIO with perturbation covariance $10^{-4}$ $rad^2$, SE2LIO$^\sharp$ refers to the method that does not consider the out-of-SE(2) perturbations, and SE2LO does not include IMU measurements. It can be seen that the proposed method achieved better results compared to the other methods.

\section{Conclusion}\label{section:Conclusion} 
In this paper, we propose a hybrid LiDAR-inertial SLAM framework for performing real-time state estimation and mapping in complex environments for ground vehicles. This system uses a non-iterative two-stage distortion compensation to reduce the computational cost. Moreover, we propose SE(2) constraints that involve the edge and plane features from LiDAR measurements as well as incorporating the out-of-SE(2) perturbations into the integrated noise term. The results show that SE2LIO outperforms the existing SOTA LiDAR-based SLAM systems like LOAM and FLAOM with less computational cost. We also develop a tightly coupled LiDAR-IMU fusion algorithm to generate IMU constraints between LiDAR frames. The results show that our proposed system, SE2LIO, is real-time capable and delivers superior localization and mapping accuracy over the SOTA LiDAR-only method and loosely coupled methods.

Although we evaluate the proposed method on different datasets to validate the robustness, the proposed method may fail when the slope of the ground varies significantly. The future work can include more experiments to study the impact of applying SE(2) constraints on different uneven ground planes, such as lawns and beaches. Furthermore, we plan to integrate the SE(2) constraints on both LiDAR and IMU factors in our future work. 


\bibliographystyle{IEEEtran}
\bibliography{IEEEabrv.bib}

\end{document}